\documentclass{article}
\usepackage{arxiv}

\usepackage[utf8]{inputenc} 
\usepackage[T1]{fontenc}    
\usepackage{hyperref}       
\usepackage{url}            
\usepackage{booktabs}       
\usepackage{amsfonts}       
\usepackage{nicefrac}       
\usepackage{microtype}      
\usepackage{lipsum}		
\usepackage{graphicx}
\usepackage{amsmath}
\usepackage{algorithm}
\usepackage{algorithmic}
\usepackage{subfigure}

\title{GL-GAN: Adaptive Global and Local Bilevel Optimization model of Image Generation}

\author{ \href{https://orcid.org/0000-0000-0000-0000}
	{\includegraphics[scale=0.06]{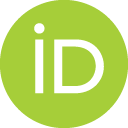}\hspace{1mm}Ying Liu}\\
	College of Informatics\\
	Huazhong Agricultural University\\
	Wuhan, 430070, China\\ 
	\And
	\href{https://orcid.org/0000-0000-0000-0000}{\includegraphics[scale=0.06]{orcid.png}\hspace{1mm} Wenhong Cai} \\
	College of Informatics\\
	Huazhong Agricultural University\\
	Wuhan, 430070, China \\	
	\And
	\href{https://orcid.org/0000-0000-0000-0000}{\includegraphics[scale=0.06]{orcid.png}\hspace{1mm}Xiaohui Yuan} \\
	College of Informatics\\
	Huazhong Agricultural University\\
	Wuhan, 430070, China \\	
	\And
	\href{https://orcid.org/0000-0000-0000-0000}{\includegraphics[scale=0.06]{orcid.png}\hspace{1mm}Jinhai Xiang} \\
	College of Informatics\\
	Huazhong Agricultural University\\
	Wuhan, 430070, China \\	
}

\hypersetup{
pdftitle={A template for the arxiv style},
pdfsubject={q-bio.NC, q-bio.QM},
pdfauthor={David S.~Hippocampus, Elias D.~Striatum},
pdfkeywords={First keyword, Second keyword, More},
}

\begin{document}
\maketitle

\begin{abstract}
	Although Generative Adversarial Networks have shown remarkable performance in image generation, there are some challenges in image realism and convergence speed. The results of some models display the imbalances of quality within a generated image, in which some defective parts appear compared with other regions. Different from general single global optimization methods, we introduce an adaptive global and local bilevel optimization model(GL-GAN). The model achieves the generation of high-resolution images in a complementary and promoting way, where global optimization is to optimize the whole images and local is only to optimize the low-quality areas. With a simple network structure, GL-GAN is allowed to effectively avoid the nature of imbalance by local bilevel optimization, which is accomplished by first locating low-quality areas and then optimizing them. Moreover, by using feature map cues from discriminator output, we propose the adaptive local and global optimization method(Ada-OP) for specific implementation and find that it boosts the convergence speed. Compared with the current GAN methods, our model has shown impressive performance on CelebA, CelebA-HQ and LSUN datasets.
\end{abstract}


\section{Introduction}
Generative adversarial networks~\cite{1} are a powerful class of image generation compared with VAE~\cite{5} and flow models. Notably, the results show an excellent performance by designing new model architectures and the implementation of stability techniques. However, by the studying to some models, we observe that many synthetic images show quality imbalance performance within a  sample. In short, the other areas of the generated image show impressive performance except for some small range poorly portions. For example, some models~\cite{16,17} excel at synthesizing images with global structure (e.g., image outline, the location of eyes and the hairstyle in face), which is prone to achieve by images' global optimization; while it fails to focus on some details, such as artifact, distorted and uncoordinated regions, which always appears in some images more or less. One possible explanation is that the common global optimization models may be to ignore some small areas due to its fairly low loss proportion. Therefore, it is possible that small range low-quality areas within an image have not been optimized during training. In practice, this in turn also can be a reason why some early GAN models ~\cite{2,3,4,9} with simple structure only can generate relatively low-quality images. Some models well leverage the structural superiority to pay more attention to small low-quality areas by increasing structural complexity but it is at the cost of low computational efficiency.

\begin{figure*}
	\includegraphics[width=16cm]{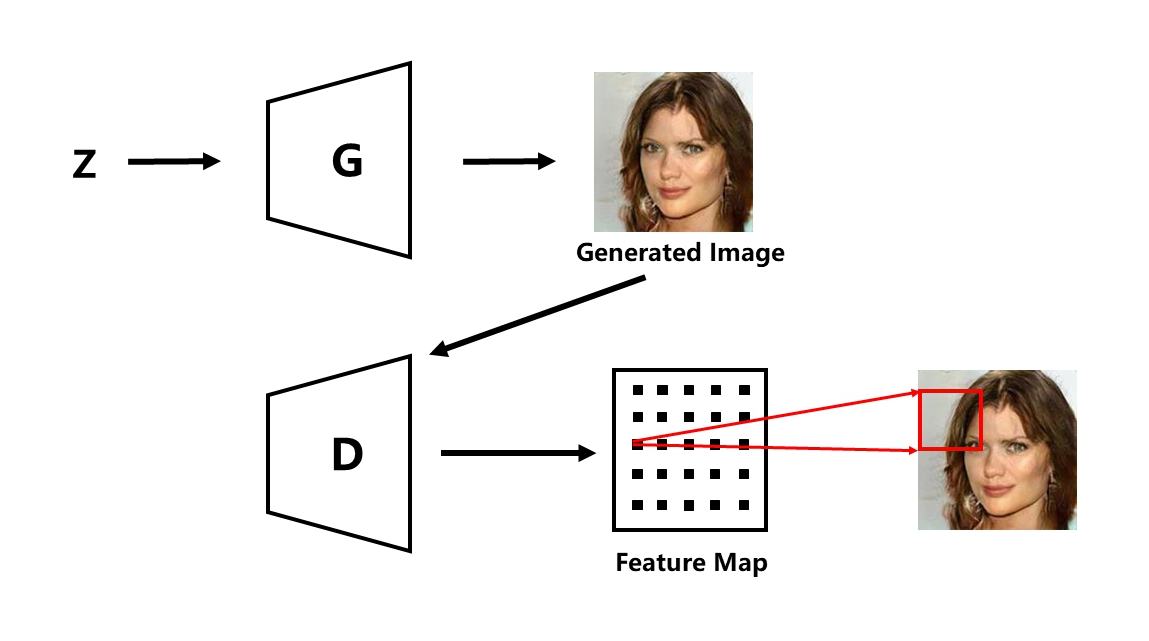}
	\caption{Architecture of GL-GAN, where the red box is the receptive field corresponding to the element in feature map.}
	\label{fig:one}
\end{figure*}

\begin{figure*}
	\centering
	\includegraphics[width=16cm]{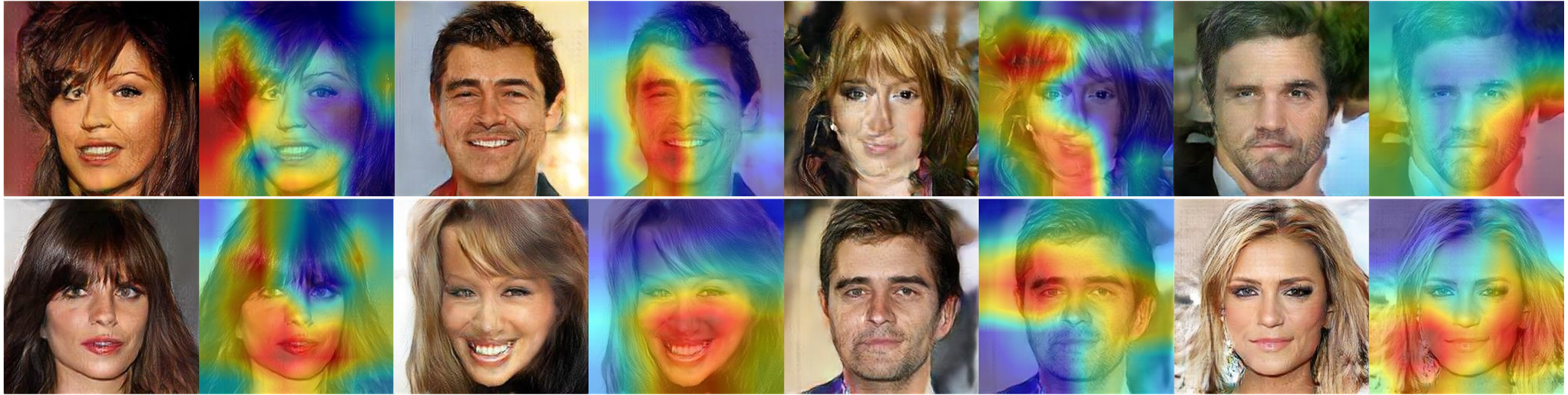}
	\caption{Location about  low-quality areas of generated images on CelebA-HQ256 dataset. The color from blue to red indicates that the quality of region is from good to bad.}
	\label{fig:two}
\end{figure*}

\begin{figure*}
	\centering
	\includegraphics[width=16cm]{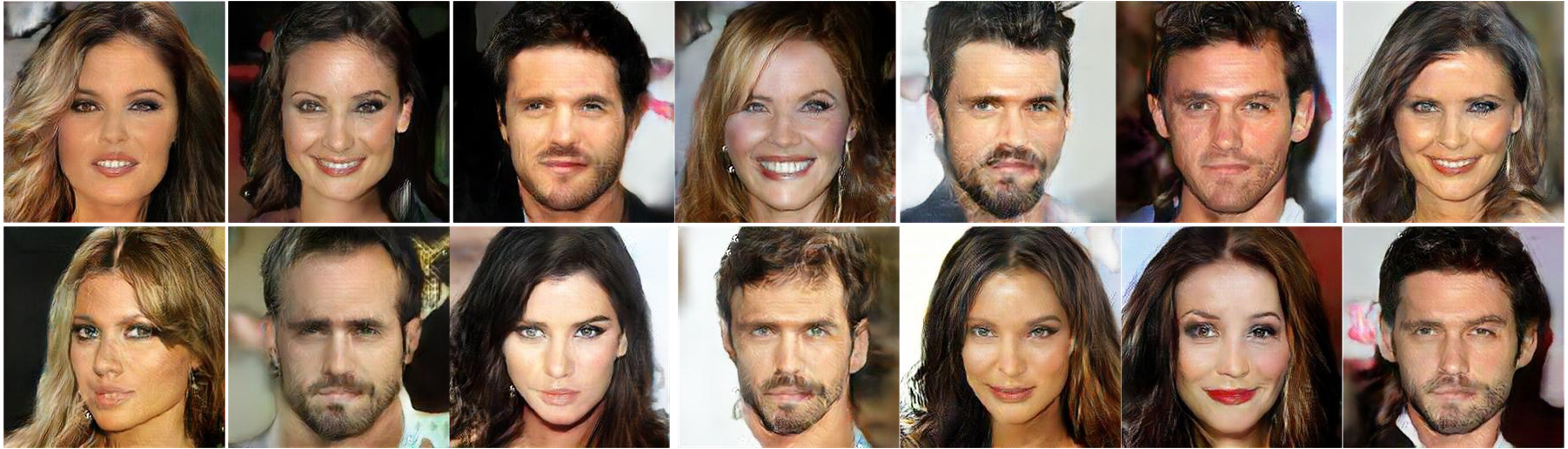}
	\caption{Generated images on CelebA-HQ256 dataset by GL-GAN.}
	\label{fig:three}
\end{figure*}

In this work, we propose an adaptively global and local bilevel optimization model(GL-GAN), which applies the local bilevel optimization model with a traditional global optimization model. According to the feature map of discriminator output, we obtain the quality measure of each area in an image.  Then the local bilevel optimization model can be regarded as a reliable guiding strategy for generator optimization by accurately capturing and optimizing the low-quality areas of a sample. To stabilize training, we investigate the spectral normalization and  apply the local-norm to our model. 

In addition, we conduct the adaptively global and local bilevel optimization method(Ada-OP) based GL-GAN theory which coordinately optimizes the whole and local of the image. In this work, the standard deviation between elements (loc-std) within the same feature map is used as an index to measure local quality balance nature in an image. We argue the mean of elements within a feature map as the whole quality measure. Then the quality standard deviation between images is regarded as global quality merit(glo-std). When the glo-std is greater than a certain value (this indicates that there is a wide range unbalance area), the global optimization is performed, otherwise, the local bilevel optimization is enforced.

Extensive experiments that have been conducted on CelebA, CelebA-HQ(see Fig \ref{fig:three}) and LSUN datasets validate the effectiveness of the proposed model in the generation of high-quality images. During applying the local bilevel optimization model, only the low-quality region is optimized thus it causes the reduction in the optimization space, which helps with speeding up the convergence speed and improving the computational efficiency.

\section{Related Work}
{\bf Generative Adversarial Networks} GANs~\cite{1} have become a central part of generation tasks compared with other models. Notably, GANs show impressive performance in sample quality by various types of methods, including designing new net architecture~\cite{2,11,12}, modifying the loss function~\cite{7} and adding conditional restriction~\cite{9,10}. Recently, BigGAN~\cite{26} which adapts the orthogonal regularization method and timely truncates the input prior distribution z greatly improves GAN's generating performance. Progressive GAN~\cite{27} and Style-GAN~\cite{28} adapt progressive growth method to train GAN through layer-by-layer to generate high-resolution images.

In addition, there are also many progresses to maintain the stability of training. The main purpose is to stabilize training by ensuring models' Lipschitz continuity, which motivates the development of weight clipping\cite{3}, gradient penalty\cite{4} and spectral normalization\cite{6}. Meanwhile, WGAN-QC~\cite{29} proposes an optimal transport regulator (OTR) based on the theory of secondary transport cost to stabilize training. Through the analysis of Dirac-GAN, ~\cite{8} shows the necessity of absolute continuity for convergence.

{\bf Feature Map} Feature map has been a very concerned concept that can capture a specific feature(e.g.style, figure, location) in an image. Several early models mainly apply the feature map information to achieve style transfer, which is presented as a loss function~\cite{13,14}. StarGAN~\cite{17} and DRPAN~\cite{15} both take the feature map information as an area's quality measure in a sample, which helps with generating high-quality details. Nevertheless, there are some limits in models, including applying the feature map of small size and large computational cost.
SAGAN~\cite{25} learns to efficiently find global, long-range dependencies within the feature map of images through the self-attention mechanism. In this work, we regard the local information within the feature map as a basis to process the adaptively global and local bilevel optimization.

\section{Adaptive Global and Local Bilevel Optimization Model}

Most GAN-based models aim at measuring the whole image quality by conduct the global optimization method, which is mainly implemented by the output probability of discriminator. Global optimization mode roughly focuses on the quality of the overall area by a output value, thus it is nontrivial for some small details within an image to fine modify. In this section, we propose an adaptive global and local bilevel optimization model, which can guide the generator to reasonably modify parameters based on the max-min two-player game of GAN.

\subsection{Feature Map}
By analyzing some samples generated by GANs models, we observe that there are always some small low-quality areas within an image. To explicitly illustrate this phenomenon, the hot maps about low-quality areas are presented(see Fig \ref{fig:two}), where the red areas denote the low-quality regions. Therefore, we argue that the quality distribution within a synthetic image is unbalanced.

Considering using the output of discriminator as the whole image quality metric in some models, we regard that using the output obtained by a patch model to represent a receptive field quality metric within an image is also reasonable. The patch model inherits from the idea of PatchGAN in ~\cite{16}. To capture the small range low-quality details, we establish a discriminator model, whose output is a feature map.

\begin{equation}
\begin{aligned}
y_{h\times w}=D_{\varphi}(x)
\label{eq:one}
\end{aligned}
\end{equation}

Where $D_{\varphi}$ represents the discriminator with parameter $\varphi$ and $x$ is a sample. The output of discriminator $y_{h\times w}{\in}{R^{h\times w}}$ is a matrix, where every element in the matrix corresponds to a receptive field of an image. Specifically, $y_{i,j}$, which is an element in matrix $y_{h\times w}$, denotes the quality evaluation of the i-th row and the j-th column receptive field.

The adaptive global and local bilevel optimization model is to optimize the generator parameters in the global and local perspectives. In order to carry out the method, we first build the assessment model of receptive field level for discriminator:

\begin{equation}
\max\limits_{\varphi}E_{x_{\sim P_{data}}}[f(D_{\varphi}(x))]+E_{z\sim{P_z},G_{\theta}(z)\sim{P_g}}[f(-D_{\varphi}(G_{\theta}(z)))]
\label{eq:two}
\end{equation}

Where $P_{data},P_{g}$ respectively denote the distribution of real samples and generated samples.
$G_{\theta}$ denotes the generator with parameter ${\theta}$.
$f:R^d\longrightarrow R$  is an operation function to the output of the discriminator, such as the sum function, the mean function, the linear function and the nonlinear function, which is utilized to get various types of loss functions, where we choose the hinge loss among all the optimization formulas.
The model target is the same as the original model, which is to distinguish the generated images from the real images.

\subsection{Local Bilevel Optimization Model}

Based on the feature map mentioned in Section 3.1, we construct a local bilevel optimization model, which optimizes the local low-quality areas of the generated image by the bilevel method.
The low-resolution regions are first captured by dot multiplying the output of discriminator with a mask matrix $h^*$, where the mask matrix is obtained by inner level optimization and composed of 0 and 1. Then the generator ($G_{\theta}$) optimizes the low-quality areas by using gradient descent algorithm.

\begin{equation}
\begin{aligned}
\mathop
{Object1=\min\limits_{\theta}}&E_{z\sim P_z}[f(h^*\odot(-D_{\varphi}(G_{\theta}(z))))] \cr
&s.t.\ h^*\in arg \max\limits_{h\in H} \sum_{h,w}{(h\odot (D_{\varphi}(G_{\theta}(z))-\alpha))}
\label{eq:three}
\end{aligned}
\end{equation}

The objective of inner layer in (\ref{eq:three}) is to select an optimal mask matrix $h^*$ when other parameters $\theta,\varphi$ are fixed. To simplify the choice of mask matrix $h$, we empirically design a constant $\alpha$ as the criteria for evaluating quality (ideally, when the values from the output are lower than the standard, the values in mask matrix corresponding to the values are 1, and vice versa the values are 0). Where $H=\{h_1,h_2,\cdots,h_n,\cdots\}$ is a matrix set, in which each matrix has the same size with the discriminator's output and is made of $0$ and $1$. The $\odot$ denotes the dot product.

The objective of the outer layer is to optimize the generator parameters ${\theta}$ directionally based on the unrealistic receptive fields, which is selected by the optimal mask matrix based on the internal optimization.

\subsection{Adaptive Global and Local Optimization Method(Ada-OP)}
Adaptive global and local optimization method generates high-resolution images by adaptively conducting global optimization (which focuses on the whole image as the optimizing objective) and local optimization (which only optimizes the low-quality areas within an image) during training generator. The local optimization model has been shown in Section 3.2. To clarify the method, we first build the global optimization model about the generator.

\begin{equation}
\begin{aligned}
\mathop
{Object2=\min\limits_{\theta}}E_{z\sim P_z}[f((-D_{\varphi}(G_{\theta}(z))))] 
\label{eq:four}
\end{aligned}
\end{equation}

In the training process, different extent quality differences between receptive fields and the quality  differences between images both will affect the selection of optimization mode. So it is necessary to define some metrics that can measure global or local differences in quality. We choose the mean(${\bar\sigma}$)(in a batch of images) of the standard deviation between different receptive fields within an image to measure the local differences. And the standard deviation($\sigma$) between different images is selected to measure the global difference, where we regard the mean($\mu_k$) between different receptive fields within an image as the whole image quality evaluation.

The global standard deviation $\sigma$ can be used as the standard to measure whether it needs to conduct the global optimization or local optimization. Because if $\sigma$ is high, there is a large range of quality differences between images, so global optimization should be carried out. 
The intuition that we do this is:
rough image first is generated (the overall quality should be basically the same), then the details are optimized. The evaluation metric for global optimization is as follows:

\begin{equation}
\begin{aligned}
&\mathop
{\mu_{k}}=\frac{\sum_{i,j}y_{i,j}}{h\cdot w} ,{\mu}=\frac{\sum_{k=1}^K\mu_k}{K} \cr
&{\sigma}=\frac{\sum_{k=1}^k(\mu_k-\mu)^2}{K},k\in\{1,2,\cdots,K\}
\label{eq:five}
\end{aligned}
\end{equation}

Where $K$ is the batch-size of images and $\mu_k$ represents the quality of the k-th image. $\mu$ denotes the quality mean of all $K$ images. When $\sigma\ge\beta$ where $\beta$ is a constant, global optimization is executed, otherwise local optimization is carried out.

When implementing local optimization, the choice of the mask matrix depends on the local size of unrealistic regions. Thus we divide the level of local bilevel optimization mode into $I,II$ and $III$, and the corresponding constant $\alpha$ in formula (\ref{eq:three}) are respectively $\alpha_1,\alpha_2,\alpha_3$. The higher level ($III$) defines the larger local optimization scope, so the constant $\alpha$ in formula (\ref{eq:three}) is bigger. The specific local standard deviation is as follows:

\begin{equation}
\begin{aligned}
&\mathop
{\sigma_{k}}=\frac{\sum_{i,j}(y_{i,j}-\mu_{k})^2}{h\cdot w}\cr
&{\bar\sigma}=\frac{\sum_{k=1}^K\sigma_{k}}{K},where\ k\in\{1,2,\cdots,K\}
\label{eq:six}
\end{aligned}
\end{equation}

Where $\sigma_k$ is the evaluation standard deviation of all receptive fields in the k-th image. Owing to the mean of the different standard deviation corresponding to different local scope, the larger $\bar\sigma$  means larger internal difference , so the level is higher.

On the basis of the above definition, the specific local and global optimization method can be given as follows:

\begin{equation}
\begin{aligned}
\mathop
{Objective=}\begin{cases}
Object2 & \text{if}\ \sigma\ge\beta \\
Object1,where\  \alpha=\alpha_1  & \text{if}\  \bar \sigma\le{\delta_1} \\
Object1,where\  \alpha=\alpha_2 & \text{if}\  {\delta_1} \le\bar\sigma\le{\delta_2} \\
Object1,where\  \alpha=\alpha_3 & \text{if}\  \bar \sigma\ge{\delta_2}
\end{cases}
\label{eq:seven}
\end{aligned}
\end{equation}

Where $\delta_1,\delta_2$ are both constants to divide the local standard deviation into different limits. Based on the method, the generator performs adaptive global and local bilevel optimization. Finally, we give the algorithm1 to clarify the training process. 

\begin{algorithm}[tbh]
	\caption{GL-GAN}
	\label{alg:algorithm}
	\begin{tabular}{l}
		\textbf{Input}: Real data $X$, batch-size $m$, epoch $n,k_G$. Adam parameters,\\ $\alpha,\beta_1,\beta_2$ \\
		\textbf{Output}: $G_{\theta},D_{\varphi}$ 
	\end{tabular}
	\begin{algorithmic}[1]
		\FOR{$i=0$ to $n$}
		\STATE Sample $\{x_i\}_{i\in I}\sim P_{data}$ for real data.\\
		\STATE Sample $\{z_j\}_{j\in J}\sim P_z$ random noise.\\
		\STATE Let $y_j=G_{\theta}(z_j),\forall j\in J$.\\
		\STATE $g_{\varphi}\gets\ the\  gradient\ of\ (\ref{eq:two})$.\\
		\STATE ${\varphi}\gets Adam (g_{\varphi},\varphi,\alpha,\beta_1,\beta_2)$.\\
		\FOR{$t=0$ to $k_G$}
		\STATE $output=D_{\varphi} (G_{\theta}(z_j))$.\\
		\STATE Compute $\sigma,\bar\sigma$ according to Eq.(\ref{eq:five}),(\ref{eq:six}).\\
		\STATE Applying the global and local bilevel optimization model according to Eq.(\ref{eq:seven})
		\STATE $g_{\theta}\gets\ the\ gradient\ of\ (\ref{eq:seven})$.\\
		\STATE${\theta}\gets Adam(g_{\theta},\theta,\alpha,\beta_1,\beta_2)$.
		\ENDFOR
		\ENDFOR
	\end{algorithmic}
\end{algorithm}

\section{Experiments}

To evaluate the effectiveness of the proposed GL-GAN, we conduct extensive experiments on CelebA, CelebA-HQ and LSUN church datasets. 

In Section $4.1$, we firstly present experimental datasets and the evaluation metric. Model structure and some parameter settings are shown in Section $4.2$. Next, the feature map method and spectral normalization is investigated for capturing low-quality regions and stabilizing training in Section $4.3$. Finally, the adaptive global and local bilevel optimal method (Ada-OP) is studied and the comparison about running-times and quality with the state-of-the-art methods is presented in Section $4.4$.

\subsection{Datasets and Evaluative Criteria}
$\textbf{CelebA}$ CelebA is a large face properties dataset with $202,599$ celebrity images. In the dataset, each image has $40$ attribute annotations and $5$ landmark locations, which can be used for attribute editing and face detection. And the size of the images is all $178\times 218$. In this paper, we resize the size to $128\times 128$ for training.

$\textbf{Celeba-HQ}$ It is a high-resolution face dataset which is obtained based on CelebA. In our model, we choose the size of  $256\times256$ and $512\times512$ face images as training set. Each resolution has $30K$ images.

$\textbf{LSUN}$ The dataset is a high-resolution images dataset of $10$ scenarios, which includes bedroom, bridge, church, living-room scenes $et\ al$. The church dataset is used for training in our method. There were $7,907$ images in the dataset (the sum of training set and testing set), which  is cropped and resized to $256\times 256$ by bicubic interpolation.

$\textbf{Evaluative Criteria}$ We choose Frechet Inception Distance (FID) for quantitative evaluation. The distance between the real images and the synthetic images at the feature level is calculated as a measure. So the smaller FID means the higher quality. FID shows the more consistent with human evaluation in the realism and variation of the generated images. 

\begin{figure*}
	\centering
	\subfigure[no-norm]{
		\includegraphics[width=4.5cm]{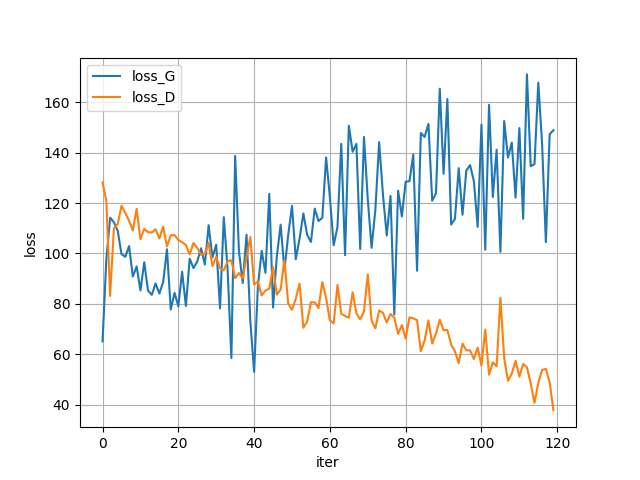}
		\label{fig:spn}}
	\hspace{0.01in}
	\subfigure[local-norm]{
		\includegraphics[width=4.5cm]{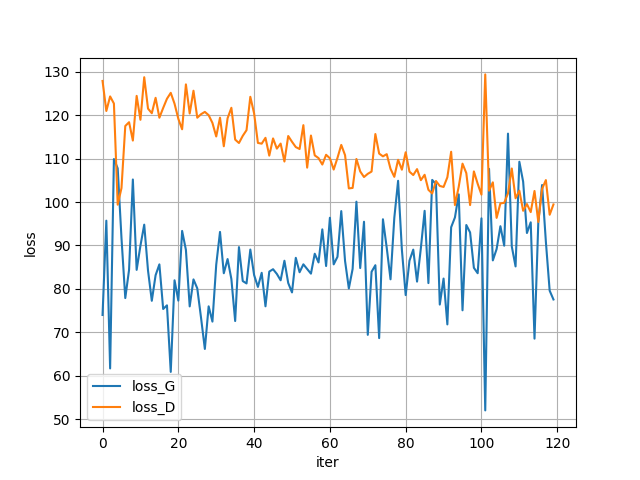}
		\label{fig:spl}}
	\hspace{0.01in}
	\subfigure[global-norm]{
		\includegraphics[width=4.5cm]{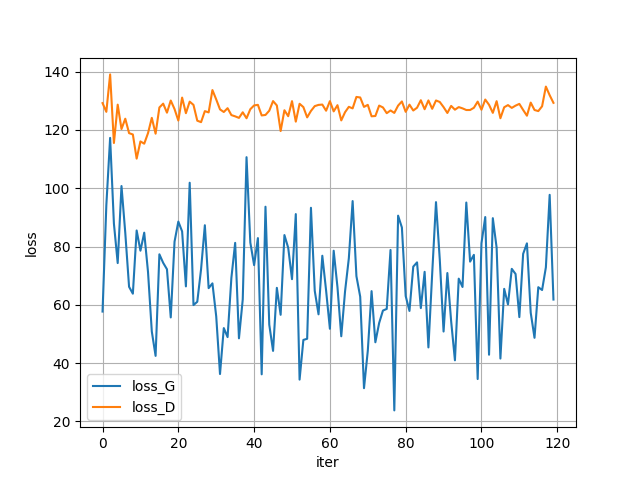}
		\label{fig:spw}}
	\caption{The loss curves about applying spectrum normalization of the different degree on the CelebA dataset.}
	\label{fig:four}
\end{figure*}

\begin{table}
	\centering
	\caption{FID scores for generative images on CelebA datasets with different degree of spectrum normalization on the baseline+patch8 method.}
	\begin{tabular}{lccc}
		\toprule
		Method & no-norm & local-norm &global-norm\\
		\midrule
		FID    &  40.01     &  14.09     &  171.02\\
		\bottomrule	
	\end{tabular}
	\label{tab:one}
\end{table}

\begin{table}
	\centering
	\caption{The generated images' FID (at the same time) on CelebA, CelebA-HQ256 and LSUN-church dataset  under different cases .}
	\begin{tabular}{lccc}
		\toprule
		Method  & CelsbA & CelebA-HQ256 & LSUN\\
		\midrule
		baseline       & 21.24  & 35.75 &  56.59   \\
		baseline+path4       & 16.67  & 20.63  &  51.29   \\
		baseline+path8    & 14.09  & 28.96    & 49.70 \\
		baseline+path4+Ada-OP  & 14.87   &  21.61   & 38.66   \\
		baseline+path8+Ada-OP   & 12.86  & 20.90 & 41.14  \\
		\bottomrule
	\end{tabular}
	\label{tab:two}
\end{table} 

\subsection{ Model Structure and Parameters}

The model of \textbf{GL-GAN} is put forward on the basis of \textbf{DCGAN}. The model is  effective to boost convergence and improve image performance with a simple structure. Generator($G$) is composed of a series of basic unit, whose structure is ConvTranspose-BatchNorm-ReLU. The input is a random vector of size $100\times1$ from the standard normal distribution. Discriminator($D$) is the same as the normal classifier with Convolution-LeakyReLU as a basic unit layer. The output is a feature map of the size of $8\times8$ or $4\times4$. Meanwhile, the parameters of the generator and discriminator respectively are carried out whole and local spectral normalization to stabilize training.
\begin{table}
	\centering
	\caption{Running time and FID comparison between different models in CelebA datasets.}
	\begin{tabular}{lcc}  
		\toprule
		Method  & FID & running-time \\
		\midrule
		WGAN-GP   &20.31    & 0.45 days    \\
		WGAN-real   & 25.23  & 0.50 days      \\
		SAGAN      &18.65 & -          \\
		WGAN-QC   & 12.9  & 0.6 days      \\
		GL-GAN    & 12.86  &   0.51 days     \\
		\bottomrule
	\end{tabular}
	\label{tab:three}
\end{table}

Except for the WGAN-QP model (which is trained on the NVIDIA TITAN Xp), all experiments are conducted on the same NVIDIA TESLA V100. 
GL-GAN adapts the Hinge loss function and Adam optimization method with $\beta_1=0.5$ and $\beta_2=0.999$. By default, the learning rate for generator is 0.003 and the learning rate for discriminator is 0.0008. The learning rate is kept unchanged during training with 1:1 balanced updates for the discriminator and generator. Except the CelebA dataset’s batch-size is $64$, the rest is all of $16$. In addition to $30$ epochs trained on CelebA, the rest are trained with $70$ epochs.

\subsection{Spectral Normalization and Feature Map}

In this section, we analyze the effectiveness of local spectral normalization (local-norm) and feature map method in improving the stability and finding out the low-quality regions.

In terms of stability, we find that the method of executing global and local spectrum normalization to generator and discriminator respectively(called local-norm) is more effective. In Fig \ref{fig:four} and Tab \ref{tab:one}, we both present the results by applying different degree spectral norm methods with model, where no-norm, local-norm and global-norm respectively represent no, local and global implementation of spectrum normalization.   As can be seen from the Fig \ref{fig:four}(a), the loss curve about D has great change and about G is fairly unstable without spectral normalization. The reason could be that it is not enough to stabilize the training only by regularization techniques. In contrast, the loss curve about D hardly changes in the condition of the global spectral norm(see Fig \ref{fig:four}(c)). It indicates that using the global spectral(about G and D) norm will lead to the strict parameter limitation, which has a negative effort to convergence speed. Compared with no-norm and global-norm method, the curve of the local-norm implies that the model realizes the mutual progress between generator and discriminator with stable loss change. One possible explanation is that local-norm method has a certain limitation on parameters and is beneficial to the stable training. Besides, we also show the FID scores of images about implementing the three methods in models in Tab \ref{tab:one}, which can see that the FID score (14.09) obtained by local-norm is the lowest. These results both demonstrate that applying local-norm with model is effective to stabilize training and further improve the generation quality. The baseline mentioned later in the paper refers to the method of adapting local-norm on the original model.

Owing to the appearance that some local regions have poor generation quality within some generated images compared with other regions, the feature map, which is as the output of discriminator, is applied to the model to further select the low-quality regions. It can be seen from the Fig \ref{fig:two} that the output's feature map can accurately find the low-quality regions (i.e., the red region), which implies that it is reasonable to using feature map as a basis for detail modification. We experiment on CelebA, CelebA-HQ and LSUN datasets under different conditions respectively(i.e., the baseline model, baseline+patch4 and baseline+patch8). As can be seen from the Tab \ref{tab:two}, the baseline+patch(4 or 8) method has a lower FID score compared with the baseline method in all the three datasets, where baseline, patch4, patch8 and Ada-OP respectively denote using the local-norm method in original models, using the size of $4\times4$ feature map, using the size of $8\times8$ feature map and using the adaptive global and local bilevel optimization method. The low FID scores of the feature map method on three datasets show that it is general to improve the synthetic image performance.

\begin{figure}
	\centering
	\subfigure[baseline+patch8$\times$8]{
		\includegraphics[width=6cm]{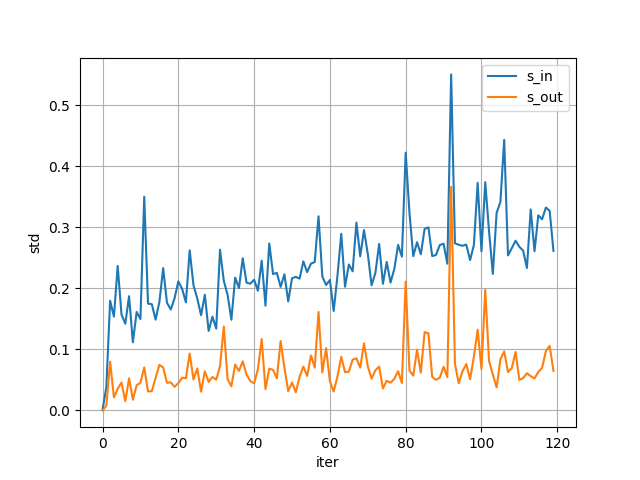}
		\label{fig:sta}}
	\hspace{0in}
	\subfigure[baseline+patch8$\times$8+Ada-OP]{
		\includegraphics[width=6cm]{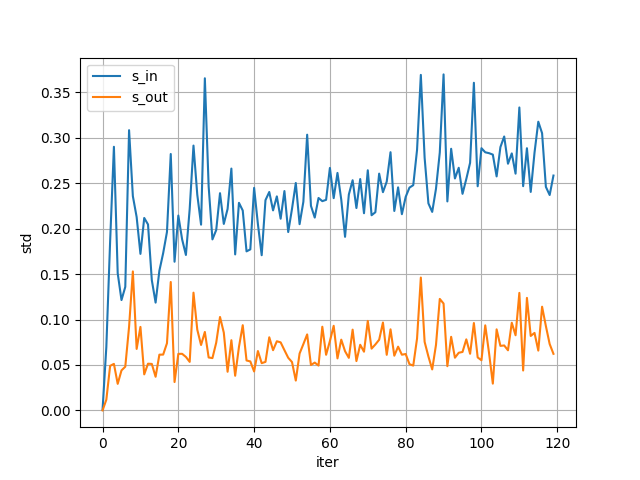}
		\label{fig:stb}}
	\caption{Standard deviation changes curves in different training modes about CelebA.}
	\label{fig:eight}
\end{figure}

\begin{figure*}
	\centering
	\subfigure[baseline+patch8$\times$8]{
		\includegraphics[width=12cm]{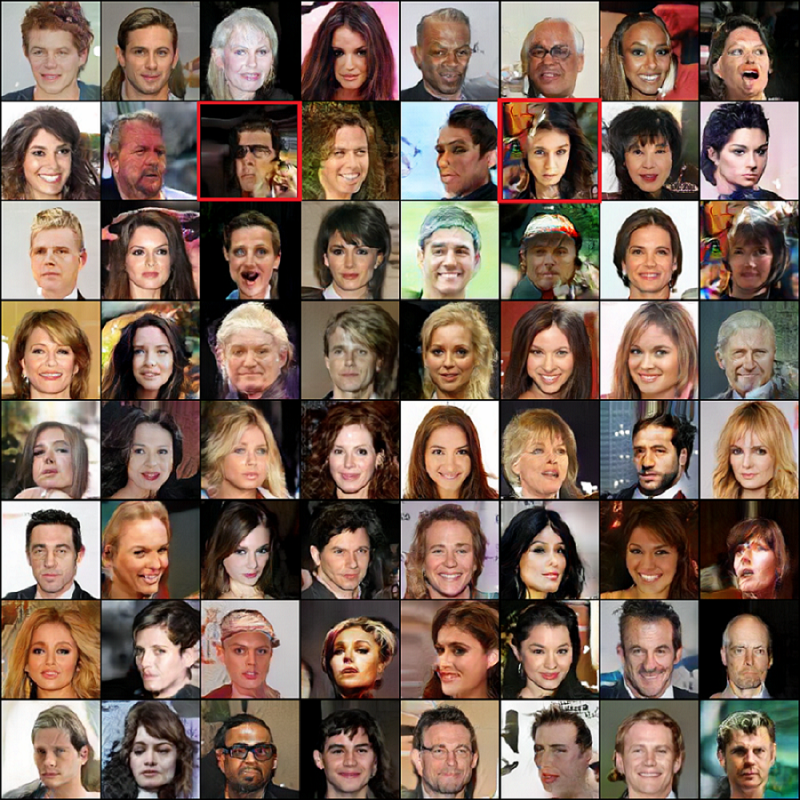}
		\label{fig:imga}}
	\hspace{0.01in}
	\subfigure[baseline+patch8$\times$8+Ada-OP]{
		\includegraphics[width=12cm]{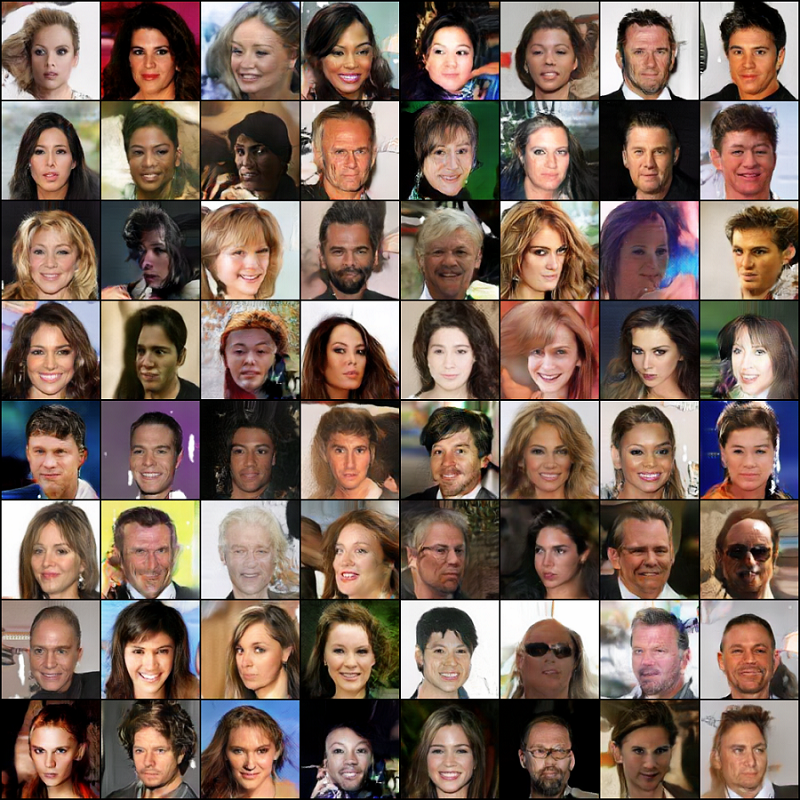}
		\label{fig:imgb}}
	\label{fig:five}
	\caption{Generated images in different training modes about CelebA dataset.}
\end{figure*}

\begin{figure*}
	\centering
	\includegraphics[width=16cm]{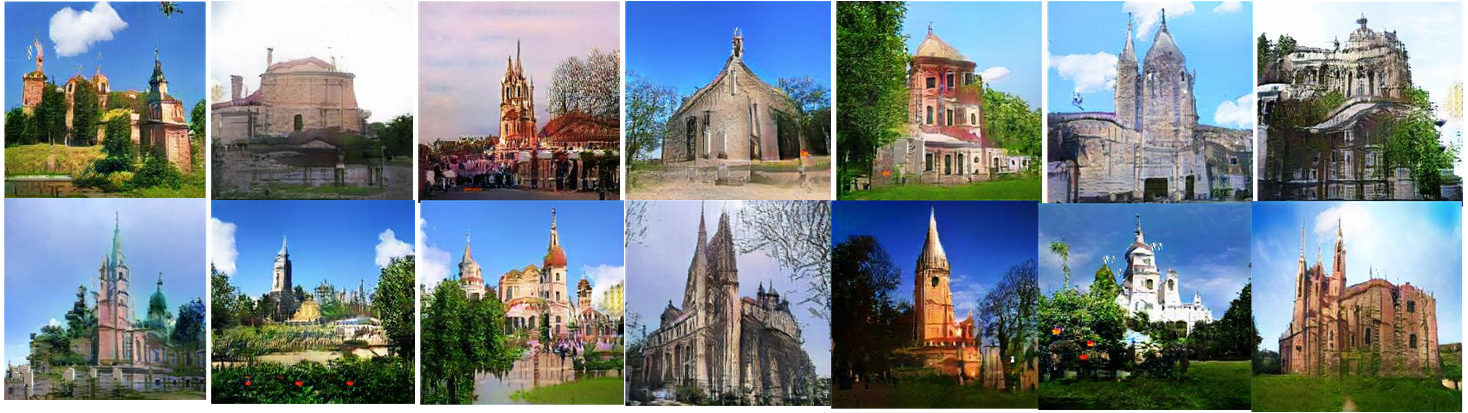}
	\caption{Randomly generated high-resolution images by GL-GAN method on LSUN church dataset.}
	\label{fig:six}
\end{figure*}

\begin{figure*}
	\centering
	\includegraphics[width=16cm]{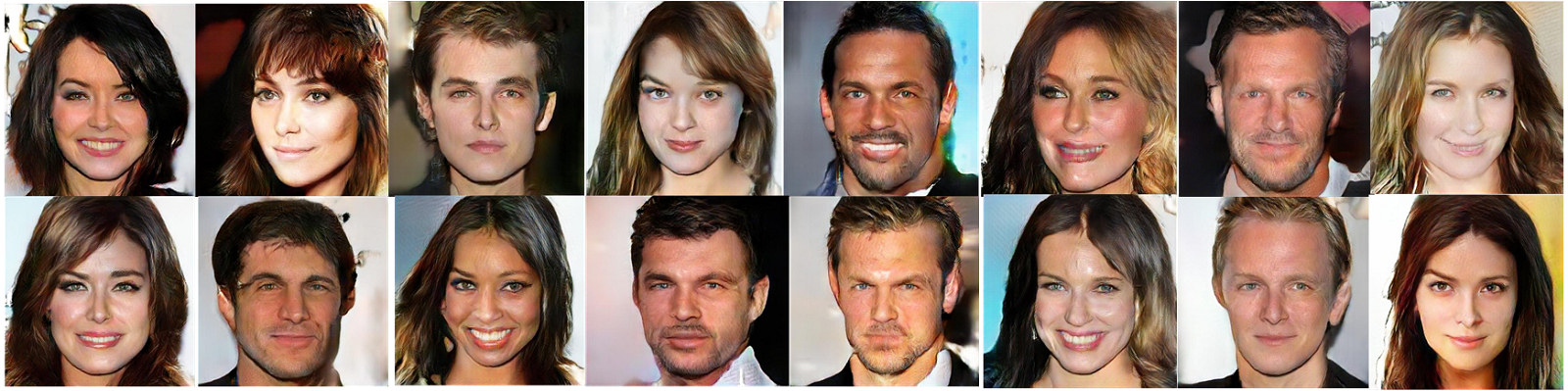}
	\caption{Randomly generated high-resolution images by GL-GAN method on Celeba-HQ512 dataset.}
	\label{fig:seven}
\end{figure*}

\subsection{Ada-OP and comparison with the state-of-the-art}

Based on the feature map obtained by discriminator, we carry out adaptive global and local bilevel optimization method(Ada-OP) according to local and global standard deviation of the feature map (Please refer to Formula (\ref{eq:seven}) for the specific optimization method). To test the performance of the training method, we show the change curves of standard deviation under the situation of applying or not applying the method. Compared to Fig \ref{fig:sta} (without applying the method), the change range of two standard deviations is both smaller and more stable with Ada-OP (Fig \ref{fig:stb}). In addition, we can observe that the generated images look more realistic with fewer artifacts in Fig \ref{fig:imgb} than the images in Fig \ref{fig:imga} (the artifact images marked with red boxes). From Tab \ref{tab:two}, we also can see that it can achieve the optimal FID with baseline+patch+Adapt optimization method in all three datasets. These results both demonstrate that Ada-OP is effective to reduce the inner difference and accurately modify low-quality regions.

The baseline+patch+Ada-OP method is adopted as the final optimization model(GL-GAN)in the paper. The images randomly generated with the model on LSUN church and CelebA-HQ512 datasets are shown in Fig \ref{fig:six} and Fig \ref{fig:seven}. From the perspective of qualitative analysis, it can be seen that the GL-GAN can achieve a relatively outstanding effect on high-resolution images.
Compared to the baseline, FID scores decrease by 15 points on the CelebA-HQ256 in Tab \ref{tab:two}.

In addition, the implementation of GL-GAN is beneficial to improving the convergence speed of the model and reducing the running time. The Tab \ref{tab:three} lists the FID score and training times for the CelebA dataset in different models (WGAN-GP, WGAN-real, SAGAN, WGAN-QC, GL-GAN), where WGAN-QC is trained on the NVIDIA TITAN Xp(it has higher operating efficiency than NVIDIA TENSLA V100), and other models are trained on the NVIDIA TENSLA V100. In almost the same amount of time, our method produces the higher-realistic images, which is lower 7 percentage points than WGAN-GP in FID. This indirectly shows that our method can boost the convergence owing to the reduction of computation during local optimization.

\section{Conclusion}

The paper proposes an adaptive global and local bilevel optimization model based on  an imbalance of quality distribution within the generated image. The feature map from the discriminator's output is adapted to locate the low-quality areas. In order to realize the local optimization, we construct a \textbf{local bilevel optimization model}, which can select and optimize the poor quality areas. We achieve adaptively guided optimization about global and local of generated images by adaptive global and local bilevel optimization method (\textbf{Ada-OP}). On the dataset CelebA, CelebA-HQ and LSUN, the convergence speed and the quality of image both have an excellent improvement.

In our method, the local and global optimization method has room for improvement. For example, only the low-quality rectangular receptive field can be filtered out in our model. And there may be an overlap between rectangular receptive fields. In future work, we will figure out the more appropriate local and global optimization method and further study the model stability.


\bibliographystyle{unsrt}
\bibliography{references}  






\end{document}